\documentclass[runningheads]{llncs}
\usepackage{colortbl}
\usepackage{multirow}
\usepackage{graphicx}
\usepackage{caption,subcaption}
\begin{document}
\title{Analysing the Predictivity of Features to Characterise the Search Space\thanks{This work is supported by TUBITAK, Turkey, with sponsoring for Dr Durgut's research visit hosted by UWE Bristol, UK.}}
\author{
    Rafet Durgut\inst{1}\and%\orcidID{0000-0002-6891-5851} \and
    Mehmet Emin Aydin\inst{2}\and%\orcidID{0000-0002-4890-5648} \and
    Hisham Ihshaish\inst{2}\and%\orcidID{0000-0001-5530-4894} \and
    Abdur Rakib\inst{2,3}%\orcidID{0000-0001-5430-450X}
}
\authorrunning{R. Durgut et al.}
\institute{Bandirma Onyedi Eylul University, Bandirma, Turkey \\
\email{rdurgut@bandirma.edu.tr}\\\and
University of the West of England, Bristol, UK\\
\email{\{mehmet.aydin,hisham.ihshaish,rakib.abdur\}@uwe.ac.uk}\\ \and
Centre for Future Transport and Cities, Coventry University, Coventry, UK\\
\email{rakib.abdur@coventry.ac.uk}
}
\maketitle              % typeset the header of the contribution
\begin{abstract}
Exploring search spaces is one of the most unpredictable challenges that has attracted the interest of researchers for decades. One way to handle unpredictability is to characterise the search spaces and take actions accordingly. A well-characterised search space can assist in mapping the problem states to a set of operators for generating new problem states. In this paper, a landscape analysis-based set of features has been analysed using the most renown machine learning approaches to determine the optimal feature set. However, in order to deal with problem complexity and induce commonality for transferring experience across domains, the selection of the most representative features remains crucial. The proposed approach analyses the predictivity of a set of features in order to determine the best categorization.

%Exploring search spaces is one of the most unpredictable challenges that has attracted the interest of researchers for decades. One way to handle unpredictability is to characterise search spaces and take action, accordingly. A reasonably well-characterised search space may help map the problem states to the set of operators so as to produce the next problem states. In this paper, landscape analysis-based set of features have been analysed to identify the best feature set using machine learning techniques. The selection of the most representative features, however, remains critical in order to deal with problem complexity and to induce commonality for transferring experience cross domains. This paper analyses the predictivity of a set of features, in this respect, to move towards the best characterisation.

%%AR revised above%%Exploring through search spaces is one of very un-predictive problems, which attracts researchers attention for decades. One approach to characterise the search spaces would be using features inspiring of machine learning problems. However, in order to handle problem complexity, selection of the most representative features remains crucial. In this paper, we analyse the predictivity of a set of features to move towards the best characterisation.  

\keywords{Feature analysis  \and search space characterisation \and supervised machine learning.}
\end{abstract}
\section{Introduction}
Optimisation is the process of searching for the best fitting solution within a solution space. Search process uses instruments to achieve moving between the neighbouring solutions by the means of neighbourhood functions, also know as operators. Operators produce new solutions, but the replacement of the produced solutions or promoting them into the recognised population of solutions retains substantial challenges. Various metaheuristic approaches instrumentalise different approaches to promote the produced solutions \cite{sotoudeh2018bibliography}. Many studies drive focus on the characteristics of search space and the fitness landscape with which more information extracted through can be used for better promotion rules and higher success rate \cite{fragata2019evolution}.   

Adaptive operator selection appears to be another useful avenue to maintain diversity and richness in the search process in order to avoid potential local optima points \cite{fialho2010adaptive}. This approach is usually applied with population-based metaheuristics, i.e., evolutionary algorithms \cite{sun2020adaptive} and swarm intelligence algorithms \cite{durgut2021adaptive}. The compelling challenge always enforces to pay more attention in the way how to build the adaptive selection scheme and which kind of information to use in opting the most suitable operators. 

Fitness landscape studies have been attractive for a long time with which more auxiliary information can be extracted and used for identification of the search and the characterisation of the search space. More details can be found in one of latest reviews \cite{fragata2019evolution}. Such auxiliary information can be utilised to harvest for representative and discriminating features to characterise the search circumstances, while, previously, the problem state has been used to help characterise the search circumstances \cite{durgut2021adaptive}\cite{durgut2022transfer},  but, the approach was not scalable for different size of problem instances due to strong dependency to the problem size. 
%it is not scalable due to that the dependency to the problem size. 
This study is expected to support to hypothesise a scalable approach through a bespoke set of features.      

The aim of this study is to pave an avenue to identify the best set of predictive features in characterising the search space and fitness landscape so as to make the most efficient decision in selecting the relevant actions such as activating the best fitting/productive neighbourhood function. Predictive analysis is expected to let us dive-down in the causal effects the behaviours of neighbourhood functions in producing the the neighbouring solutions. Details of predictive analysis have been introduced in \cite{nyce2007predictive}. 

%Teng et. all. \cite{teng2016self} introduced a comprehensive study around adaptive operator selection problem focusing on modeling with Markovien Decision Processes. 

The rest of this paper is organised as follows; Section~\ref{sec:rw} provides the relevant background and related work, while Section~\ref{sec:ls} introduces the details of fitness landscape information items used previously, and selected for use in this study including population-based and individual-based measures. Section~\ref{sec:er} includes experimental details of the relevant discussions, and Section~\ref{sec:conc} concludes and outlines future work.  %while Section~\ref{sec:conc} indicates the concluded remarks.

\section{Related Work}
\label{sec:rw}
Data-driven and bottom-up approaches -- using data analysis -- in characterisation of unknown problems have been eased and facilitated with the introduction of big-data, which escalated to dealing with huge number of data instances and features. The search spaces in optimisation domain is known as an-predictable and dynamic processes, where the search space size increases exponentially as the number of dimensions grows. Attempts to characterise such search spaces faces increasing the computational complexity of most learning algorithms - for which the number of input features and sample size are critical parameters. In order to reduce the space and computational complexities, the number of features of a given problem should be reduced \cite{durgut2020feature}. Many predictors benefit from the feature selection process since it reduces overfitting and improves accuracy, among other things~\cite{SHEKAR14}. In the literature~\cite{wang2017population,MaciasEscobar2019}, fitness landscape analysis has been shown to be an effective technique for analysing the hardness of an optimization problem by extracting its features. Here, we review some existing approaches that are most closely related to the work proposed in this paper.

In~\cite{wang2017population}, the notion of population evolvability is introduced as an extension of dynamic fitness landscape analysis. The authors assumes a population-based algorithm for sampling, two metrics are then defined for a population of solutions and a set of neighbours from one iteration of the algorithm. Because of the exploration process that occurs during each generation, population evolvability can be a very expensive operation. To avoid a computationally intensive operation, the work suggests that the number of sampled generations must be carefully defined. In\cite{MaciasEscobar2019}, a very similar approach has been proposed to apply population evolvability in a hyper-heuristic, named Dynamic Population–Evolvability  based  Multi-objective  Hyper-heuristic. In~\cite{TAN2021142}, the authors proposed a differential evolution (DE) with an adaptive mutation operator based on fitness landscape, where a random forest based on fitness landscape is implemented for an adaptive mutation operator that selects DE's mutation strategy online. Similarly, in both \cite{sallam2017landscape} and \cite{SALLAM2020113033}, DE embedded with an adaptive operator selection (AOS) mechanism based on landscape analysis for continues functional optimisation problems.  

A survey by Malan~\cite{Malan2021} summarises recent advances in landscape analysis, including a variety of novel landscape analysis approaches and studies on sampling and measure robustness. it drives attention on landscape analysis applications for complex problems and explaining algorithm behaviour, as well as algorithm performance prediction and automated algorithm configuration and selection. In~\cite{teng2016self}, the authors propose a continuous state Markov Decision Process (MDP) model to select crossover operators based on the states during evolutionary search. For AOS, they propose employing a self-organizing neural network. Unlike the Reinforcement Learning technique, which models AOS as a discrete state MDP, their neural network approach is better suited to models of AOS that have continuous states and discrete actions. However, usually MDP based model computationally expensive due to the state space explosion problem. 

The majority of these studies have considered population-based landscape metrics to characterise the situation, while some have considered individual-based measures. In this study, we attempt to use both population and individual-based metrics side-by-side and to evaluate the impact of each upon the prediction results in order to consider a wide-range of information aspects in characterisation of search space. In addition, the state-of-the-art literature implemented approaches to solve functional optimisation problems, which are significantly different from combinatorial problems with respect to predictability and characterisation of fitness landscape. We attempt to solve two combinatorial problems (binary in this case), which can be seen more un-predictable in this respect. 

%\textcolor{blue}{Difference between the existing and the proposed work to be written...}
%\textcolor{red}{One thing from Rafet: the others didn't measure the efficiency and importance level of features, they used without evaluation.}

\section{Landscape Features}
\label{sec:ls}
Fitness landscape analysis provides representative information, which can be used in characterisation of the search space and the position of the problem state in hand. A vast literature has been developed over last few decades that can be utilised in selecting the the most representative information. The relevant literature can be found in \cite{fragata2019evolution,ochoa2019recent,pitzer2012comprehensive}.

Diversity is one of very important aspects of swarms to help characterise the states \cite{erwin2020diversity}, while Wang et. all \cite{wang2017population} discuss evolvability of populations with dynamic landscape structure. 

A number of features can be retrieved from state of art literature as listed in tables below Table~\ref{tab:feature_pop} and Table~\ref{tab:feature_indvl}. The population-based metrics -- considered as feature-- are listed in Table~\ref{tab:feature_pop} with corresponding calculation details. The first 5 metrics, $\{psd, pfd, pnb, pic, pai\}$, have been collected from \cite{teng2016self} and  implemented for (i.e. adjusted to) artificial bee colony algorithm (ABC), which is one of very recently developed highly reputed swarm intelligence algorithm \cite{karaboga2014comprehensive}. The metrics calculated based on distance measure have been binarised using Hamming distance as in \cite{erwin2020diversity} in order to adjust them to binary problem solving. The metrics, $\{pcv, pcr, eap, app\}$, are introduced and proposed in \cite{wang2017population} with sound demonstration, while $atn$ is obtained from the trail index used in ABC and utilised to measure/observe the iteration-wise hardness in problem solving. In addition, $pdd$ is picked up from \cite{anescu2017fast} to calculate the distance between two farthest individuals with in a population/swarm.       

The literature includes more metrics calculated through local search procedures. However, these kind of features, i.e. metrics, have been left out due to the scope of the study. In fact, it is known that access to preliminary information on search is not easy, hence, we encompass the change in instant search in formation online decision making. 

%\label{sec:pm}
The base notation of population-based features is as follows. Let $P=\{p_i|i=0,1,...,N\}$ be the set of parent solutions and $C=\{c_i|i=0,1,...,N\}$ be the set of children solutions reproduced from $P$, where each solution has $D$ dimensions. Also, let $F^p=\{f^p_i|i=0,1,...,N\}$ be the set of parent fitness values and $F^c=\{f^c_i|i=0,1,...,N\}$ be set of children fitness values. $g_{best}$  represents the best solution has found by so far and $p_{best}$ represents the best solution in the current population. 

\begin{table}[htb!]
\caption{\label{tab:feature_pop} Population-based features}
\centering
%\scriptsize
\begin{tabular}{|l|l|}
\hline
\textbf{Feature}                      & \textbf{Formula} \\ \hline
Population Solution Diversity         & $psd = \frac{\sum^{n-1}_{i=0}{\sum^n_{j=i+1}{\|p_i-p_j\|}}}{D \frac{n(n-1)}{2}}$ \\ \hline
Population Fitness Deviation          & $pfd = \frac{\sum^{n-1}_{i=0}{\sum^n_{j=i+1}{\|f^p_i-f^p_j\|}}}{\frac{n(n-1)}{2}}$     \\ \hline
Population of new best children       & $pnb = \frac{|{c_i | f^c_i > f^p_i }|}{N} $                 \\ \hline
Proportion of new improving children  & $pic = \frac{|{c_i | f^c_i > g_{best} }|}{N} $                 \\ \hline
Proportion of amount of improvements  & $pai = \frac{{(f^c_i - f^p_i)/f^c_i | f^c_i > f^p_i }}{N} $      \\ \hline
Proportion of Convergence Velocity    & $pcv = \frac{E[max(F^c) - max(F^p)]}{max(F^p)} $                 \\ \hline
Proportion of Convergence Reliability & $pcr = \frac{E[\|x^* - x_t\| - \|x^* - x_{t+1}\|]}{D}$                 \\ \hline
Evolutionary Ability of population    & $eap = \sum_{i \in N^{*}} \frac{ \sigma(P) |f^*(P) - f(Cf_i)}{N}$   \\ \hline
Evolvability of population            & $evp = eap \times pic$                 \\ \hline
Proportion of Average Trial Number    & $atn = \frac{\sum^{n}_{i}tn_i}{N}$            \\ \hline
The diameter of population            & $pdd = max_{i,j \in \{P,C\}}{ \| p_i-c_i\|} $   \\ \hline
\end{tabular}
\end{table}

\begin{table}[!ht]
\small
\caption{\label{tab:feature_indvl}Individual solution-based features}
\centering
\begin{tabular}{|l|l|}
\hline
\textbf{Feature}                      & \textbf{Formula} \\ \hline
Distance between $g_{best}$ and parent solutions    & $idg = \frac{\|x^* - p_i\|}{D} $  \\ \hline
Distance between parent and child solutions         & $idp = \frac{\|p_i - c_i\|}{D}$ \\ \hline
Fitness gap between $g_{best}$ and child solutions  & $ifg = (f^{x^*} - f^c_i)/f^{x^*}$ \\ \hline
Fitness gap between the parent and the offspring    & $ifp = (f^c_i - f^p_i)/f^c_i $                 \\ \hline
Distance between $p_{best}$ and parent solutions    & $idb = \frac{\|p_{best} - p_i\|}{D} $  \\ \hline
Distance between $p_{worst}$ and parent solutions   & $idw =  \frac{\|p_{worst} - p_i\|}{D}$  \\ \hline
Proportion of Trial number                          & $itn = \frac{trial_i}{trial_{max}} $  \\ \hline

\end{tabular}
\end{table}

%\label{sec:im}
On the other hand, a number of metrics -- features -- can be obtained from the auxiliary information of individual solution, which seem to serve efficiently in individual-specific aspects with which the operators can act upon significantly on case basis. The individual-related features are tabulated in Table~\ref{tab:feature_indvl}, which are mostly proposed by \cite{teng2016self} except $itn$, which is introduced in this study first time. Among these features, the success rate for operator $i$ is calculated with ${osr}_i = \frac{{sc}_i }{{tc}_i} $, where $sr$ is success counter and $tc$ is total usage counter. %The rest of them are as in Table~\ref{tab:feature_indvl}.

\section{Experimental Results}
\label{sec:er}
This experimental results have been collected over multiple runs of an Artificial Bee Colony algorithm bespoke in earlier studies embedded with a pool of operators selected each time a new solution is generated randomly selecting the operators to execute. Each successful move achieved whilst the execution of the algorithm has been picked up as a successful case and labelled accordingly.  

Two well-known combinatorial optimisation problems have been considered as test-bed; One-Max \cite{goeffon2011optimal} as unimodal and Set Union Knapsack (SUKP) \cite{lin2019hybrid} as multi-model problems. The size of benchmark problems taken under consideration for One-Max and SUKP are 1000 and 500, respectively, while the maximum number of iterations are 150 and 500, respectively. 

The preliminary experimentation demonstrated that the level of hardness and complexity very much depends on the progress of search process, hence, the whole search period is divided into three phases since it is expected that the behaviour of the operators would vary significantly over the time and stage of iterations, relevant analysis is provided in upcoming subsection.     

\subsection{Feature Exploratory Analysis}
\label{sec:fd}

A set of exploratory analyses are conducted to explore both the relevance of input features as well as their relative importance to the task of operator selection — the latter is discussed further in Section \ref{sec:oc}. The tests are evaluated for each phase of the search process, separately. That is, given the set of all input features, $A$, the aim is to examine if a subset $A' \in A$ is associated with the \textit{target} success operators, corresponding to each search phase. The assumption made here is based on  whether feature membership for $A'$ is consistent, which in turn can be used to indicate the features most prevalent at predicting \textit{success} operators, per search phase, and if comparable across the two different optimisation problems.  

\begin{figure}
    %\centering
    \includegraphics[width=1.20\linewidth]{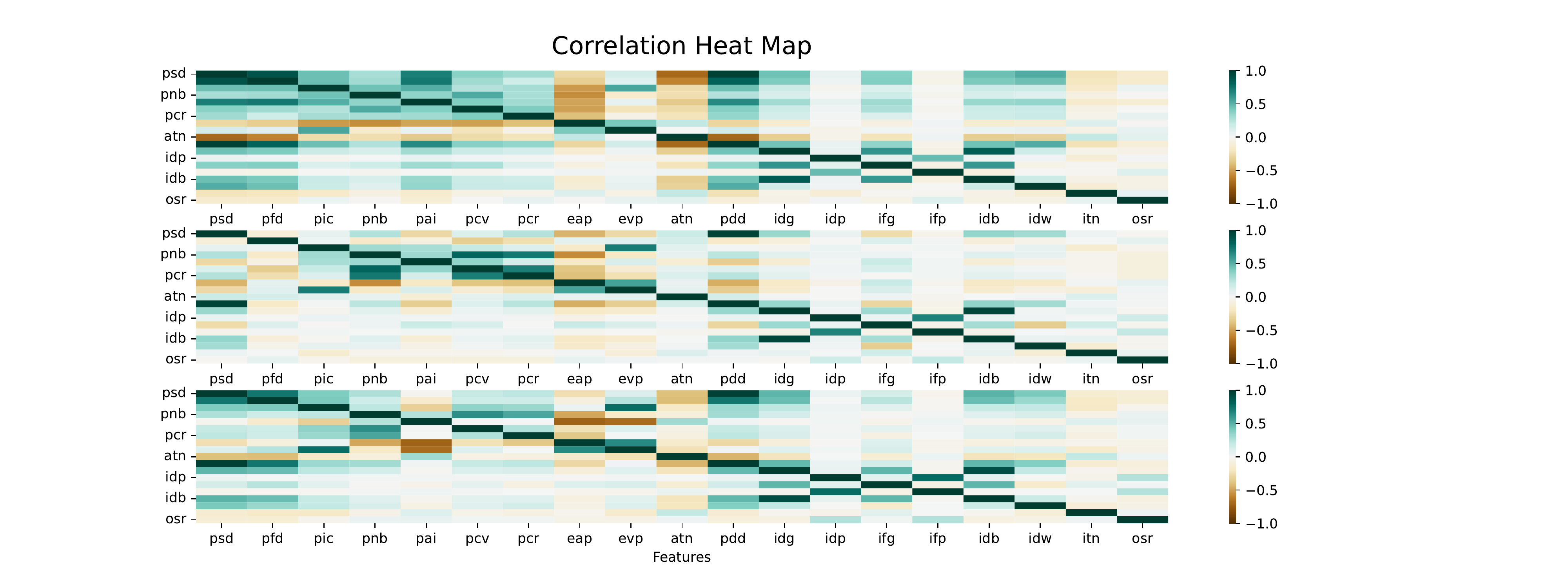}
    \caption{Pearson Correlation Coefficient matrix for the features applied to One-Max problem. The matrices are ordered top-down per search phase; 1 top and 3 down.} % (\textcolor{blue}{please check this}).}
    \label{fig:hm-fi}
\end{figure}

\begin{figure}
    %\centering
    \includegraphics[width=1.20\linewidth]{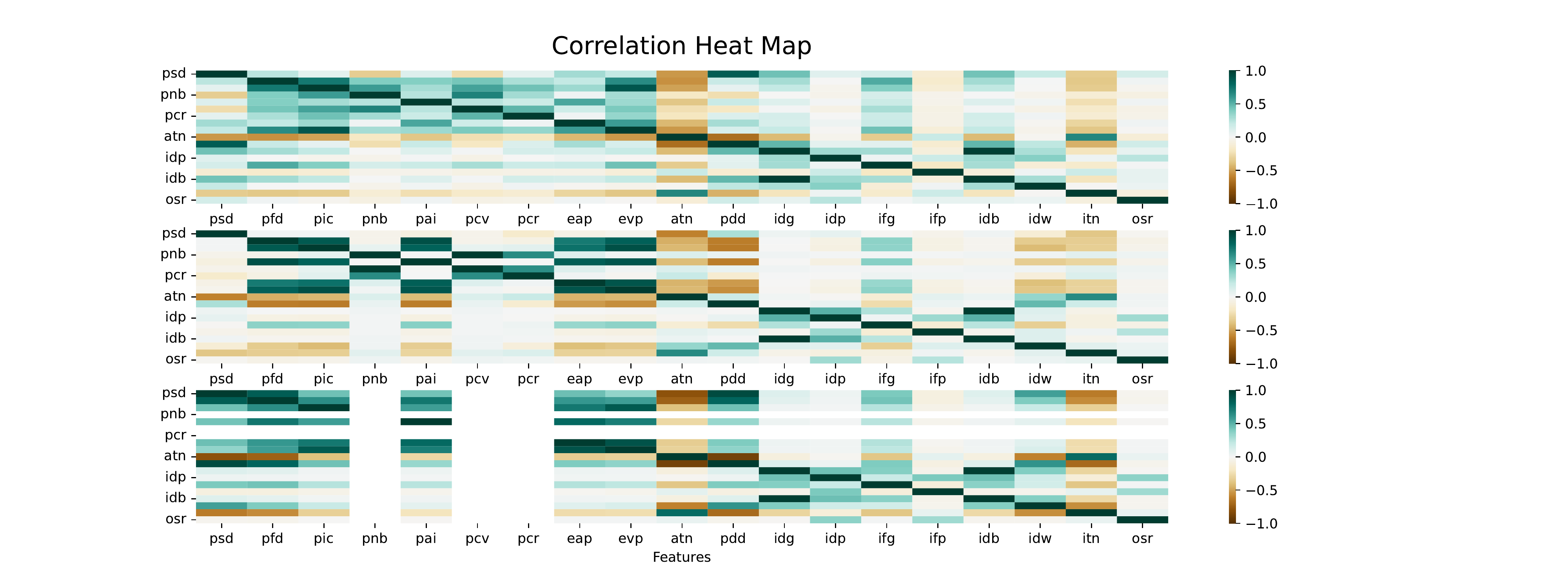}
    \caption{Pearson Correlation Coefficient matrix for the features applied to SUKP problem. The matrices are ordered top-down per search phase} % (\textcolor{blue}{please check this}).}
    \label{fig:hm-fi-sukp}
\end{figure}

The first test evaluated the strength of linear relationship between input features relative to each search phase, as shown in Fig. \ref{fig:hm-fi} for One-Max problem and in Fig. \ref{fig:hm-fi-sukp} for SUKP. 

%\label{sec:fd}
%Correlation Heatmap

There is clearly apparent linearity  -- as additionally expected, both positive as well as negative--  among different groups of features in both optimisation problems. The strength of relationship furthermore exhibits variability across the different search phases. Generally, whilst relative strength of association can be indicative for feature selection processes, further evaluation of feature importance relative to operator selection is essential, nonetheless. In particular, where membership in $A^\prime$ can be relatively stable across the two optimisation problems, we examine if the selected subset of features can learn the target variables, i.e. \textit{success} operators, associated with each problem, correctly.    

Accordingly, for both the One-Max and SUKP problems, the Chi-square ($\chi ^2$) test -- a test on whether two variables are related or independent from one another-- is conducted to examine the dependency of the response variable (\textit{success} operator) on the set of input features. $\chi^2$ statistic, computed for each feature-class pair, provides a score on the relative dependency between the values of each attribute and the different target classes. The attributes of higher values for the $\chi^2$ statistic can be said to have more importance at the task of predicting the target class, i.e. search operator, and usually as a result are selected as the input features in classification tasks. 

\begin{figure}[htb!]
  \centering
  \begin{subfigure}{\textwidth}
    \centering
    \includegraphics[width=\textwidth]{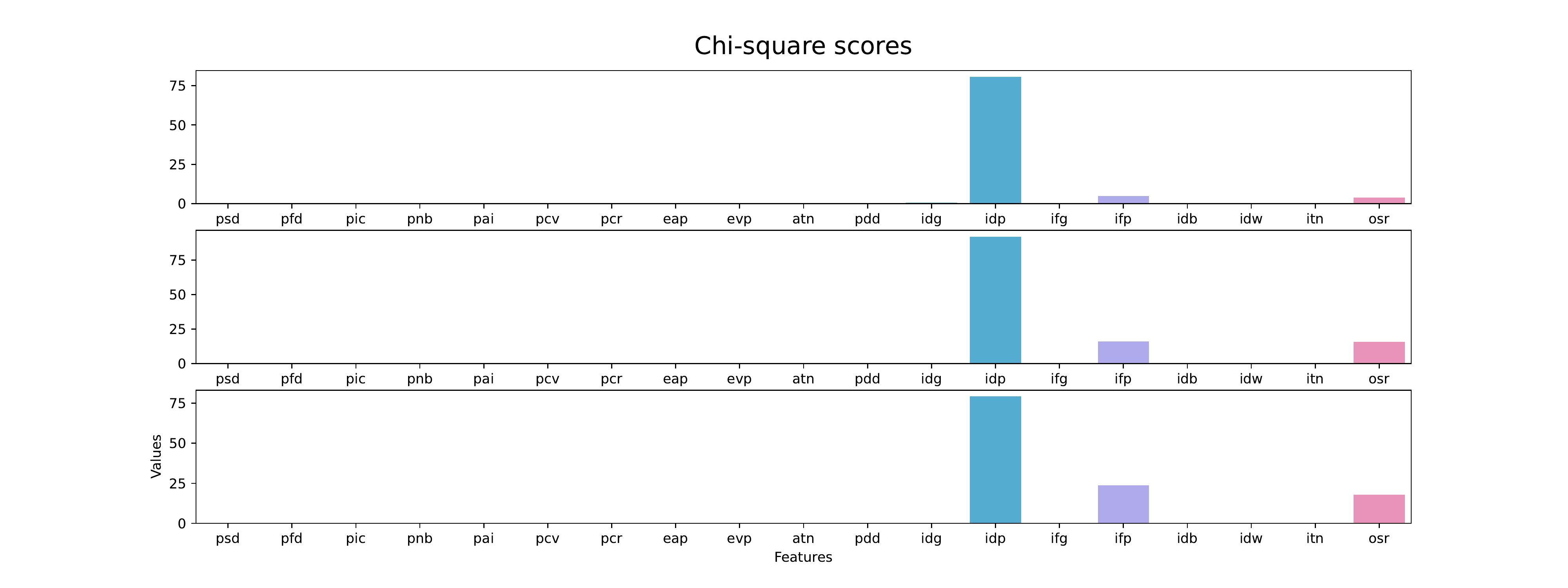}
      \caption[fig 1]{One-Max}
      \label{fig:chi2-fi}
  \end{subfigure}
  \begin{subfigure}{\textwidth}
    \centering
    \includegraphics[width=\textwidth]{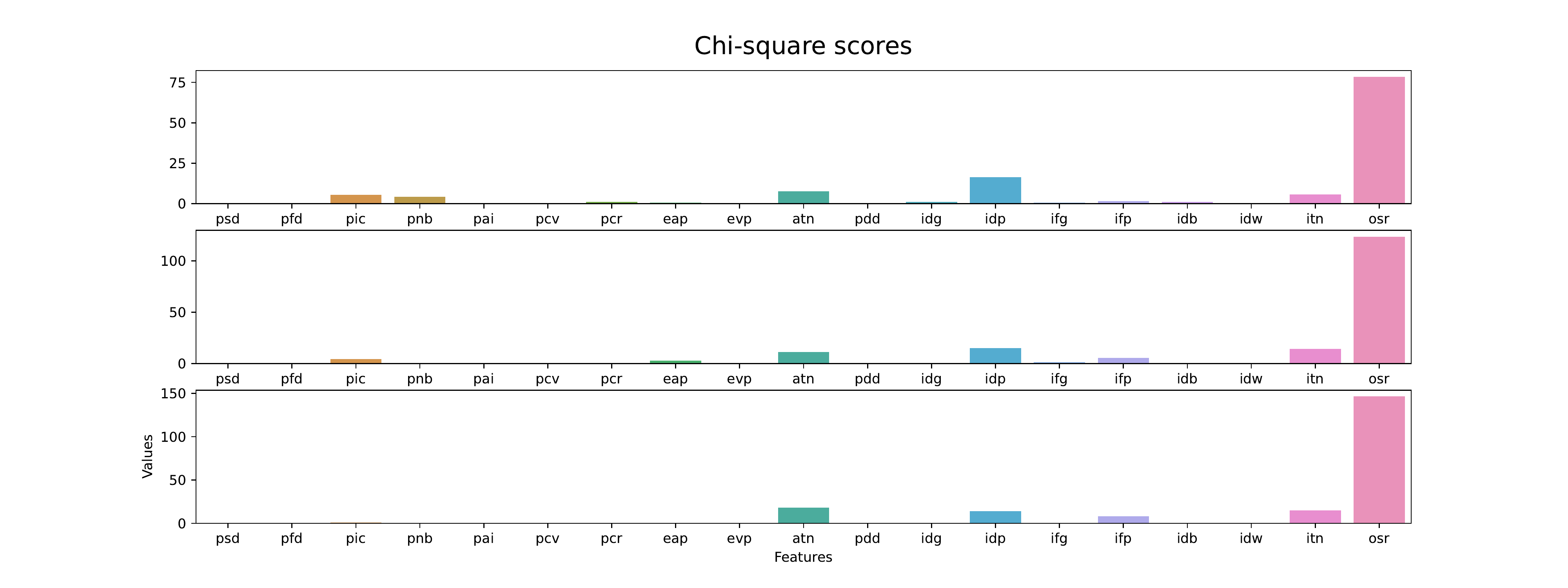}
      \caption[fig 2]{SUKP}
  \end{subfigure}
  \caption[Chi2]{Chi-square statistic rank for input features on successful search operators. Again, in both (a) and (b), ranking is ordered top-down per search phase.}
  \label{fig:chi2}
\end{figure}

The resulted ranking of input features relative to both optimisation problems is shown in Fig. \ref{fig:chi2}. Whilst these seem to exhibit differences in importance across the two problems; namely there appears to be a higher number of relevant features in SUKP compared to those in One-Max, there is nonetheless an interesting overlap between between both regarding a subset of (dominant) input features \{$idp$, $ifp$, $osr$\}, as well as an agreement on the relative irrelevance of further features to search operators. This additionally persists across the three search phases corresponding to both examined problems. Although such finding can result primitive -- not the least conclusive given the nature of the examined problems --, the resulted similarity can nonetheless be critical to examining potential prospects leading to learning a solution path (or important features) from one problem to another. 

\subsection{Operator Classification}
\label{sec:oc}
To assess the possible transferability of selected features from one search domain to another, the prediction of the different \textit{success} operators at each search phase corresponding to the two different optimisation problems is subsequently evaluated. The \textit{success} of operators relative to each search problem and phase are shown in Table \ref{tab:groundtruth}. This provides the setting for a supervised classification task in which problem features are the independent variables and the corresponding \textit{success} operators are the target class. 

\begin{table}[htb!]
\caption{Success of operators for One-Max and SUKP search problems.}
\centering
\scriptsize
\begin{tabular}{c|cccc|r}
\hline
Problem                  & Operator & Phase 1 & Phase 2 & Phase 3 & Mean      \\ 
\hline
\multirow{4}{*}{One Max} & OP 0     & 306       & 375       & 357       & 346.00       \\
                         & OP 1     & 234       & 218       & 235       & 229.00    \\
                         & OP 2     & 304       & 405       & 456       & 388.33    \\
                         & OP 3     & 323       & 328       & 316       & 322.33    \\ 
\hline
\multirow{4}{*}{SUKP}    & OP 0     & 104       & 200       & 245       & 183.00       \\
                         & OP 1     & 494       & 150       & 89        & 244.33    \\
                         & OP 2     & 1397      & 1368      & 1487      & 1,417.33  \\
                         & OP 3     & 916       & 777       & 649       & 780.67   \\
\hline
\end{tabular}
\label{tab:groundtruth}
\end{table}

Three classifiers are applied to predict the \textit{success} operators; a multilayer perceptron (MLP) with one hidden layer (feedforward ANN with 'adam' solver), Support Vector Machine (SVM) classifier with radial basis function (rbf) kernel and a Random Forest classifiers of size 200. All models have been used in classification tasks very widely for decades, and the particular choice for RF and SVM was additionally due to their ability to provide explicit feature importance ranking alongside their prediction, which we aim to utilise in the proposed hypothesis. We report the accuracy score as the prediction measure of accuracy in Table \ref{tab:accuracy}.  
\begin{table}[htb!]
\caption{The accuracy results for both problem types achieved by machine learning approaches across 3 phases }
\centering
\scriptsize
\begin{tabular}{l|ccc|ccc}
\hline
\multicolumn{1}{l|}{} & \multicolumn{3}{c|}{One Max} & \multicolumn{3}{c}{SUKP}  \\ 
\cline{2-7}
\multicolumn{1}{l|}{} & RF   & SVM  & MLP            & RF   & SVM  & MLP         \\ 
\cline{1-7}
Phase 1            & 0.79 & 0.52 & 0.70           & 0.71 & 0.62 & 0.68        \\
Phase 2            & 0.85 & 0.63 & 0.73           & 0.79 & 0.72 & 0.75        \\
Phase 3            & 0.84 & 0.65 & 0.71           & 0.83 & 0.77 & 0.80        \\ \hline
Mean                 & 0.83 & 0.60 & 0.71           & 0.77 & 0.71 & 0.74      \\ \hline
\end{tabular}
\label{tab:accuracy}
\end{table}

\begin{figure}[ht!]
  \centering
  \begin{subfigure}{\textwidth}
    \centering
    \includegraphics[width=\textwidth]{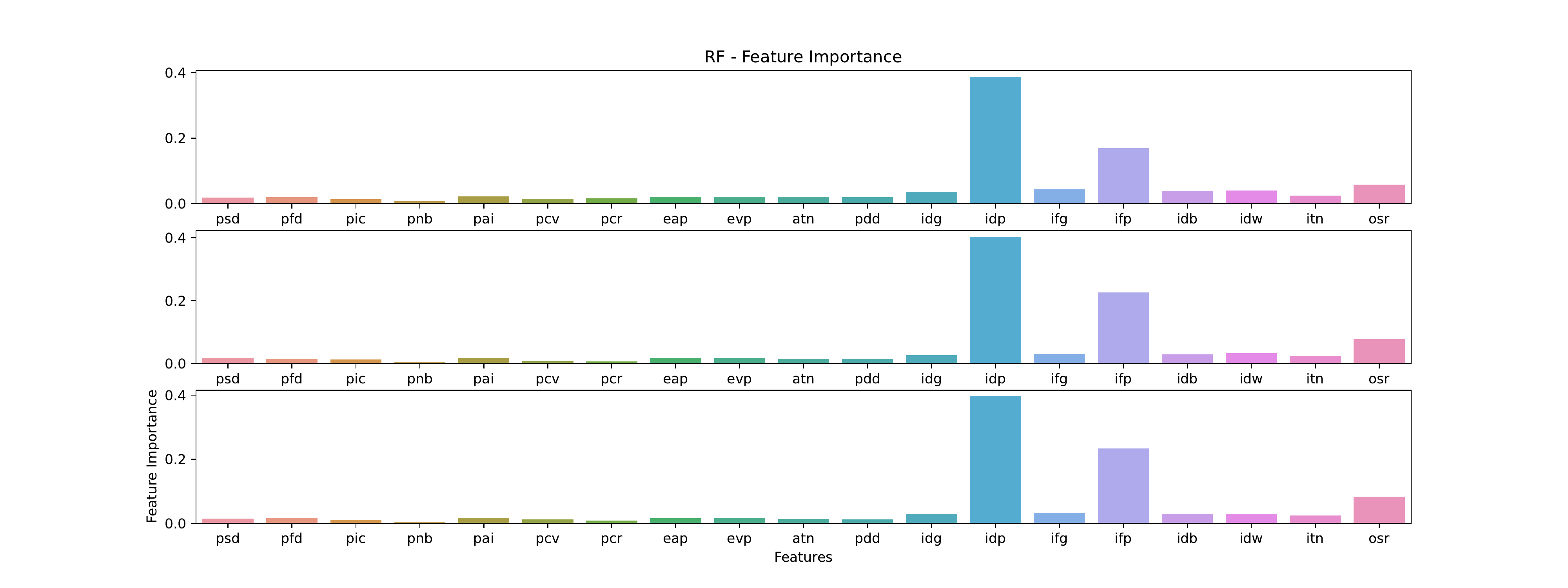}
      \caption[fig 1]{Feature importance calculated with Random Forest}
      \label{fig:rf-fi}
  \end{subfigure}
  \begin{subfigure}{\textwidth}
    \centering
    \includegraphics[width=\textwidth]{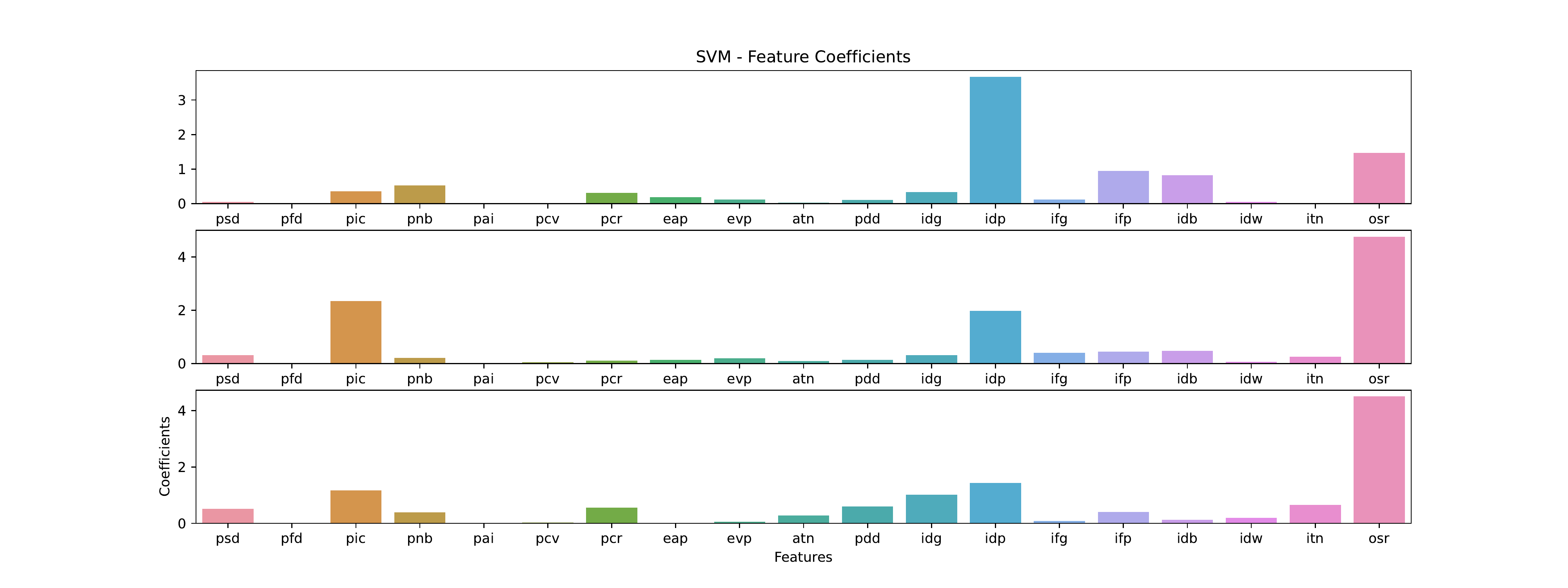}
      \caption[fig 2]{Features coefficients calculated with SVM}
  \end{subfigure}
  \caption[FI-One-Max]{Feature importance ranking for One-Max problem.}
  \label{fig:OneMac-fi}
\end{figure}

\begin{figure}[htb]
  \centering
  \begin{subfigure}{\textwidth}
    \centering
    \includegraphics[width=\textwidth]{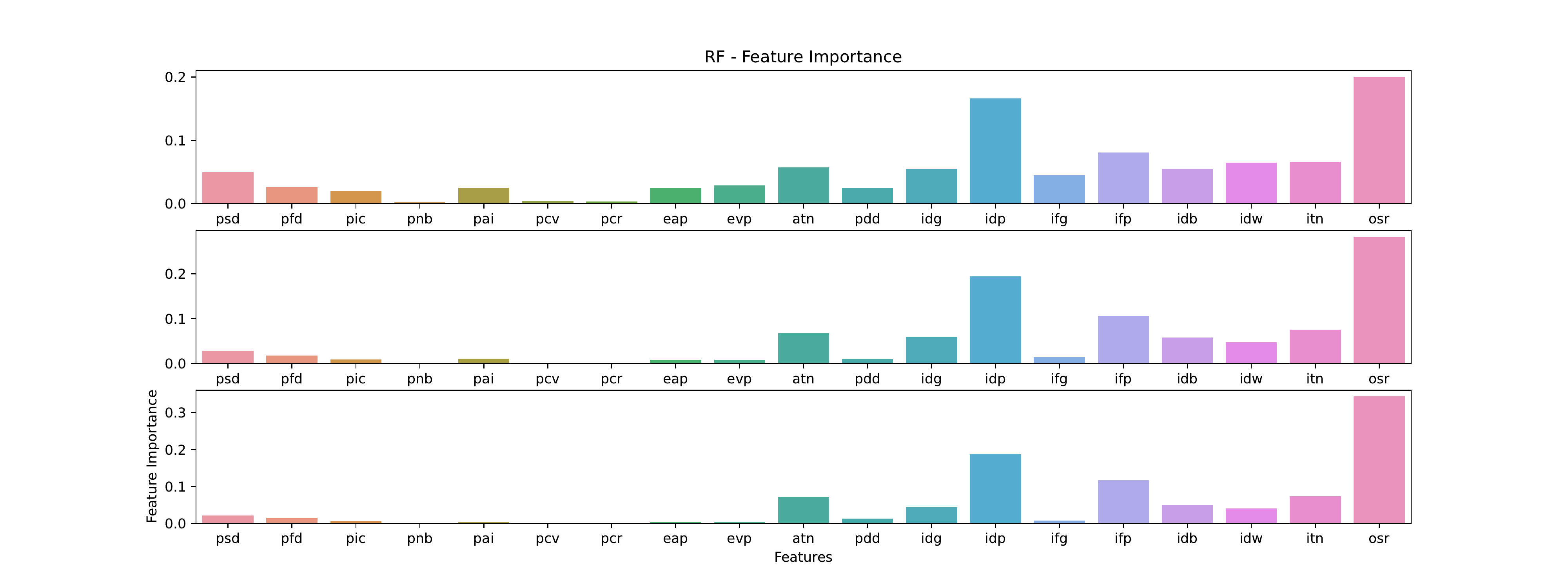}
      \caption[fig 1]{Features importance calculated with Random Forest}
      \label{fig:rf-fi-sukp}
  \end{subfigure}
  \begin{subfigure}{\textwidth}
    \centering
    \includegraphics[width=\textwidth]{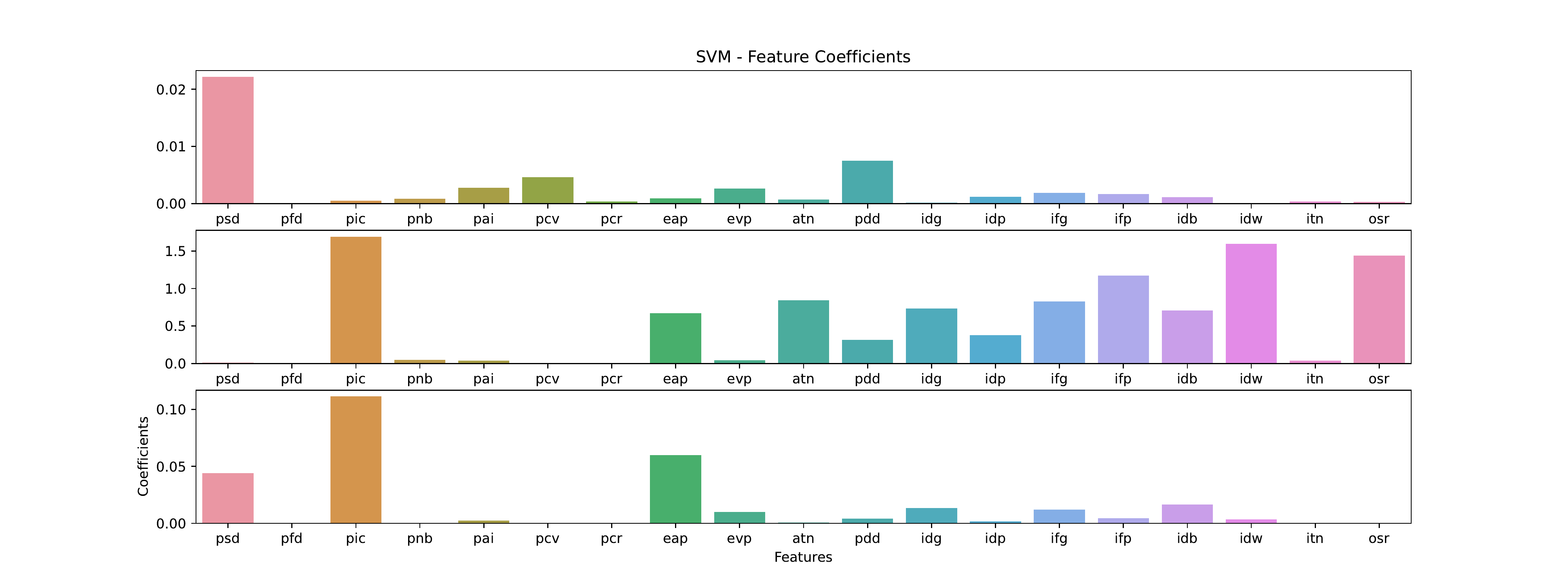}
      \caption[fig 2]{Feature coefficients calculated with SVM}
  \end{subfigure}
  \caption[FI-SUKP]{Feature importance ranking for SUKP problem.}
  \label{fig:svc-fi-sukp}
\end{figure}

Interestingly, the performance of the classifiers on both optimisation problems is relatively comparable. With the exception of SVM on One-Max which seems to be underperforming that on SUKP, the predictability of success operators from both individual as well as population domain features is consistent. It should be noted that the reported performance of the three classifiers can be tuned for further optimisation, which we aim at providing in a further study. In this study, however, the aim is to examine whether predictability of \textit{success} operators can be achieved with a subset of input features learnt in different search problem(s). In such a way the relative importance of input features for the classification tasks are computed and compared; the weighted coefficients of feature vectors in the SVM classifier as well as the importance of features from the resulted Random Forest classifier, normalised across the 200 Decision Trees between 0 and 1. The results are shown in Fig.\ref{fig:OneMac-fi} for the One-Max problem and Fig.\ref{fig:svc-fi-sukp} for SUKP.

Once again the results show promising findings as a subset of features can be seen to have similar relative importance across both search problems. In fact this emphasises the suggestion, as observed earlier in the Chi-sqaure test results, that there seems to be a subset of \textit{effective} features, like $A^\prime$, to the task of operator selection that can be transferable from one problem to another. Worth mentioning that in both Fig. \ref{fig:OneMac-fi} and Fig. \ref{fig:svc-fi-sukp}, the relative feature importance is computed for the whole set of features, as the SVM considers weighing all input attributes, and the RF calculates class impurity -- relative Shannon entropy-- weighted by the probability of reaching the target class (\textit{success} operator) corresponding to all features as these are re-sampled across 200 trees, and subsequently their scores normalised. That is to say that in selecting the subset of \textit{effective} features, their relative importance should be considered rather than the values assigned to them.

The assessment on what specific features are most prevalent to the \textit{success} operator selection, and why can be 'overenthusiastic' at this stage, especially so as this would require extensive characterisation of both search problems, which will be evaluated further in a later study. Here, however, the argument on finding a transferable $A^\prime$ from one search problem to another seems plausible. For this, the extent of predictability (solution quality) and robustness as features are reduced and transferred across different search domains should be examined further.    

%Classification Accuracy
%Confusion Matrix
\section{Conclusions and future work}
\label{sec:conc}
This paper presents an exploratory and a predictive analysis in order to reveal the impacts and domination of a set of features considered for characterisation of search spaces in optimisation domain. The idea is to identify the set of the most impactful and prominent features that best represent a problem state and its standing within its neighbourhood so that the best fitting neighbourhood function among many alternatives can be selected to generate the next problem state avoiding local optima for higher efficiency in search process.  A swarm intelligence algorithm -- artificial bee colony -- has been used with a pool of neighbourhood functions, i.e. operators, to solve two different types of combinatorial optimisation problems utilising an adaptive operator selection scheme. The set of most prominent features are elicited through a rank of weights using statistical and machine learning methods. The analysis demonstrated that a set of features mostly including individual features are found to be more discriminative than those of population-based metrics. 

The interesting preliminary outcome of the study is that the most effective features have been mostly the same even if the problem domain has changed. This can suggest that the information can be transferable between different problem domains. For the next step of this work, the success of transfer learning through the problems needs to be examined in terms of robustness and solution quality. The set features will be considered in active and reinforcement learning for dynamic and more realistic problems.

 \bibliographystyle{splncs04}
 \bibliography{ref}

\begin{thebibliography}{10}
\providecommand{\url}[1]{\texttt{#1}}
\providecommand{\urlprefix}{URL }
\providecommand{\doi}[1]{https://doi.org/#1}

\bibitem{anescu2017fast}
Anescu, G., Ulmeanu, P.: A fast self-adaptive approach to reliability
  optimization problems. Review of the Air Force Academy (2),  23--30 (2017)

\bibitem{SHEKAR14}
Chandrashekar, G., Sahin, F.: A survey on feature selection methods. Computers
  \& Electrical Engineering  \textbf{40}(1),  16--28 (2014)

\bibitem{durgut2021adaptive}
Durgut, R., Aydin, M.E.: Adaptive binary artificial bee colony algorithm.
  Applied Soft Computing  \textbf{101},  107054 (2021)

\bibitem{durgut2022transfer}
Durgut, R., Aydin, M.E., Rakib, A.: Transfer learning for operator selection: A
  reinforcement learning approach. Algorithms  \textbf{15}(1), ~24 (2022)

\bibitem{durgut2020feature}
Durgut, R., Baydilli, Y.Y., Aydin, M.E.: Feature selection with artificial bee
  colony algorithms for classifying parkinson’s diseases. In: International
  Conference on Engineering Applications of Neural Networks. pp. 338--351.
  Springer (2020)

\bibitem{erwin2020diversity}
Erwin, K., Engelbrecht, A.: Diversity measures for set-based meta-heuristics.
  In: 2020 7th International Conference on Soft Computing \& Machine
  Intelligence (ISCMI). pp. 45--50. IEEE (2020)

\bibitem{fialho2010adaptive}
Fialho, {\'A}.: Adaptive operator selection for optimization. Ph.D. thesis,
  Universit{\'e} Paris Sud-Paris XI (2010)

\bibitem{fragata2019evolution}
Fragata, I., Blanckaert, A., Louro, M.A.D., Liberles, D.A., Bank, C.: Evolution
  in the light of fitness landscape theory. Trends in Ecology \& Evolution
  \textbf{34}(1),  69--82 (2019)

\bibitem{goeffon2011optimal}
Go{\"e}ffon, A., Lardeux, F.: Optimal one-max strategy with dynamic island
  models. In: 2011 IEEE 23rd International Conference on Tools with Artificial
  Intelligence. pp. 485--488. IEEE (2011)

\bibitem{karaboga2014comprehensive}
Karaboga, D., Gorkemli, B., Ozturk, C., Karaboga, N.: A comprehensive survey:
  artificial bee colony (abc) algorithm and applications. Artificial
  Intelligence Review  \textbf{42}(1),  21--57 (2014)

\bibitem{lin2019hybrid}
Lin, G., Guan, J., Li, Z., Feng, H.: A hybrid binary particle swarm
  optimization with tabu search for the set-union knapsack problem. Expert
  Systems with Applications  \textbf{135},  201--211 (2019)

\bibitem{MaciasEscobar2019}
Macias-Escobar, T.E., Cruz-Reyes, L., Dorronsoro, B., Fraire-Huacuja, H.,
  Rangel-Valdez, N., G{\'o}mez-Santill{\'a}n, C.: Application of population
  evolvability in a hyper-heuristic for dynamic multi-objective optimization.
  Technological and Economic Development of Economy  (2019)

\bibitem{Malan2021}
Malan, K.M.: A survey of advances in landscape analysis for optimisation.
  Algorithms  \textbf{14}(2) (2021),
  \url{https://www.mdpi.com/1999-4893/14/2/40}

\bibitem{nyce2007predictive}
Nyce, C.: Predictive analytics white paper, sl: American institute for
  chartered property casualty underwriters. Insurance Institute of America p.~1
  (2007)

\bibitem{ochoa2019recent}
Ochoa, G., Malan, K.: Recent advances in fitness landscape analysis. In:
  Proceedings of the Genetic and Evolutionary Computation Conference Companion.
  pp. 1077--1094 (2019)

\bibitem{pitzer2012comprehensive}
Pitzer, E., Affenzeller, M.: A comprehensive survey on fitness landscape
  analysis. Recent Advances in Intelligent Engineering Systems pp. 161--191
  (2012)

\bibitem{sallam2017landscape}
Sallam, K.M., Elsayed, S.M., Sarker, R.A., Essam, D.L.: Landscape-based
  adaptive operator selection mechanism for differential evolution. Information
  Sciences  \textbf{418},  383--404 (2017)

\bibitem{SALLAM2020113033}
Sallam, K.M., Elsayed, S.M., Sarker, R.A., Essam, D.L.: Landscape-assisted
  multi-operator differential evolution for solving constrained optimization
  problems. Expert Systems with Applications  \textbf{162},  113033 (2020)

\bibitem{sotoudeh2018bibliography}
Sotoudeh-Anvari, A., Hafezalkotob, A.: A bibliography of metaheuristics-review
  from 2009 to 2015. International Journal Of Knowledge-based And Intelligent
  Engineering Systems  \textbf{22}(1),  83--95 (2018)

\bibitem{sun2020adaptive}
Sun, G., Yang, B., Yang, Z., Xu, G.: An adaptive differential evolution with
  combined strategy for global numerical optimization. Soft Computing
  \textbf{24}(9),  6277--6296 (2020)

\bibitem{TAN2021142}
Tan, Z., Li, K., Wang, Y.: Differential evolution with adaptive mutation
  strategy based on fitness landscape analysis. Information Sciences
  \textbf{549},  142--163 (2021)

\bibitem{teng2016self}
Teng, T.H., Handoko, S.D., Lau, H.C.: Self-organizing neural network for
  adaptive operator selection in evolutionary search. In: International
  Conference on Learning and Intelligent Optimization. Lecture Notes in
  Computer Science, vol. 10079, pp. 187--202. Springer (2016)

\bibitem{wang2017population}
Wang, M., Li, B., Zhang, G., Yao, X.: Population evolvability: Dynamic fitness
  landscape analysis for population-based metaheuristic algorithms. IEEE
  Transactions on Evolutionary Computation  \textbf{22}(4),  550--563 (2017)

\end{thebibliography}




% LNCS Homepage, \url{http://www.springer.com/lncs}. Last accessed 4
% \end{thebibliography}
\end{document}